\documentclass[10pt,journal,compsoc]{IEEEtran}
\pdfoutput=1
\ifCLASSOPTIONcompsoc
  \usepackage[nocompress]{cite}
\else
  \usepackage{cite}
\fi

\ifCLASSINFOpdf

\else

\fi

\usepackage{epsfig}
\usepackage{graphicx}
\usepackage{amsmath}
\usepackage{amssymb}
\usepackage{comment}
\usepackage{multirow}

\usepackage{booktabs}
\usepackage{array, caption, threeparttable}
\usepackage{caption}

\usepackage[table,xcdraw]{xcolor}

\usepackage{algorithm}
\usepackage{listings}
\usepackage{color}

\usepackage{mathtools}
\usepackage{todonotes}
\usepackage{gensymb}

\usepackage{hyperref}
\hypersetup{hidelinks}

\begin{document}

\title{Multi-scale Promoted Self-adjusting Correlation Learning for Facial Action Unit Detection}

\author{Xin Liu{\textsuperscript{$\ast$}},~\IEEEmembership{Senior Member,~IEEE}, 
        Kaishen Yuan{\textsuperscript{$\ast$}},  
        Xuesong Niu,~\IEEEmembership{Member,~IEEE}, \\
        Jingang Shi,~\IEEEmembership{Member,~IEEE}, 
        Zitong Yu,~\IEEEmembership{Member,~IEEE}, 
        Huanjing Yue,~\IEEEmembership{Member,~IEEE} \\
    and Jingyu Yang{\textsuperscript{$\dagger$}},~\IEEEmembership{Senior Member,~IEEE}


\thanks{Manuscript received August, 2023.}
\thanks{ $\ast$ Equal contribution. $\dagger$ Corresponding author.}

\IEEEcompsocitemizethanks{\IEEEcompsocthanksitem X. Liu is with the School of Electrical and Information Engineering, Tianjin University, Tianjin 300072, China, and also with Computer Vision and Pattern Recognition Laboratory, School of Engineering Science, Lappeenranta-Lahti University of Technology LUT, Lappeenranta 53850, Finland. E-mail: linuxsino@gmail.com

\IEEEcompsocthanksitem K. Yuan, H. Yue, J. Yang are with the School of Electrical and Information Engineering, Tianjin University, Tianjin 300072, China.

\IEEEcompsocthanksitem X. Niu is with Beijing Institute for General Artificial Intelligence, Beijing 100080, China.

\IEEEcompsocthanksitem J. Shi is with School of Software Engineering, Xi'an Jiaotong University, Xi'an 710049, China.

\IEEEcompsocthanksitem Z. Yu is with Great Bay University, Dongguan 523000, China.}

}

\markboth{IEEE Transactions on Affective Computing}%
{Shell \MakeLowercase{\textit{\emph{et al.}}}: Bare Advanced Demo of IEEEtran.cls for IEEE Computer Society Journals}

\newcommand{\etal}{\textit{et al.}}

\IEEEtitleabstractindextext{%
\begin{abstract}
Facial Action Unit (AU) detection is a crucial task in affective computing and social robotics as it helps to identify emotions expressed through facial expressions. Anatomically, there are innumerable correlations between AUs, which contain rich information and are vital for AU detection. Previous methods used fixed AU correlations based on expert experience or statistical rules on specific benchmarks, but it is challenging to comprehensively reflect complex correlations between AUs via hand-crafted settings. There are alternative methods that employ a fully connected graph to learn these dependencies exhaustively. However, these approaches can result in a computational explosion and high dependency with a large dataset. To address these challenges, this paper proposes a novel self-adjusting AU-correlation learning (SACL) method with less computation for AU detection. This method adaptively learns and updates AU correlation graphs by efficiently leveraging the characteristics of different levels of AU motion and emotion representation information extracted in different stages of the network. Moreover, this paper explores the role of multi-scale learning in correlation information extraction, and design a simple yet effective multi-scale feature learning (MSFL) method to promote better performance in AU detection. By integrating AU correlation information with multi-scale features, the proposed method obtains a more robust feature representation for the final AU detection. Extensive experiments show that the proposed method outperforms the state-of-the-art methods on widely used AU detection benchmark datasets, with only 28.7\% and 12.0\% of the parameters and FLOPs of the best method, respectively. The code for this method is available at \url{https://github.com/linuxsino/Self-adjusting-AU}.
\end{abstract}

}

\maketitle
\IEEEdisplaynontitleabstractindextext

%
\IEEEpeerreviewmaketitle

\section{Introduction}
\label{sec:intro}

\begin{figure}[t]
    \centering
    \includegraphics[width=1\linewidth]{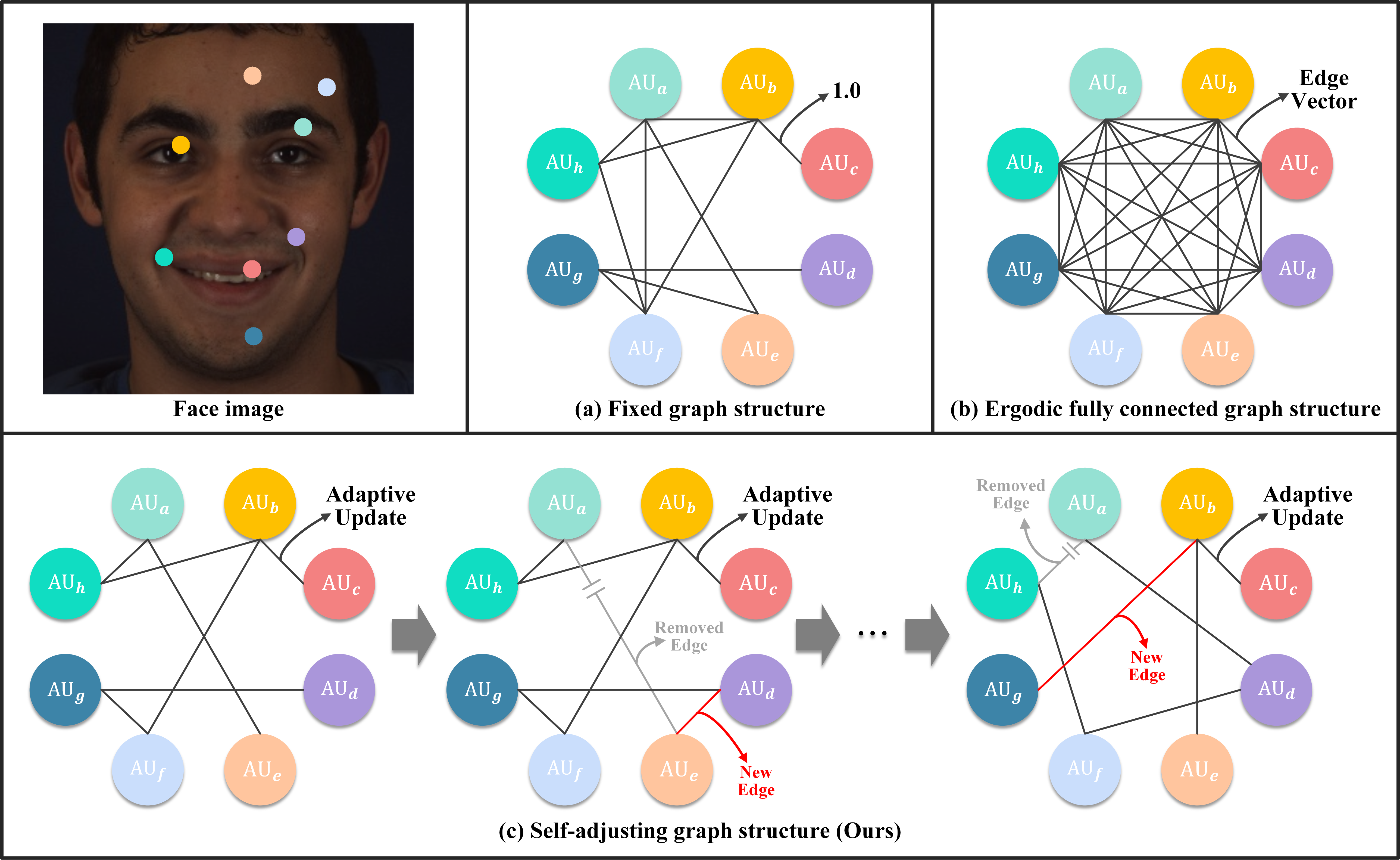}
    \caption{
    Comparison between our proposed SACL method with existing AU correlation learning methods: (a) \textbf{Fixed graph structure} based on FACS summary~\cite{chang2022knowledge} or the statistical rules of the dataset~\cite{SRERL,AU-GCN}; (b) \textbf{Ergodic fully connected graph structure} that calculates the correlation information for each pair of AUs with a high computational complexity~\cite{ME-graphAU, UGN-B}; (c) \textbf{Our self-adjusting graph structure} that constantly updates the edge connections based on the characteristics of different levels of AU motion and emotion representation information extracted in different stages to capture more implicit AU correlations with less computational effort. 
    }
    \label{fig.1}
\end{figure}

\IEEEPARstart{F}{acial} expressions are a form of body language symbols that is important for nonverbal communication. Motions of the muscles beneath the skin of the face can generate a variety of rich expressions that convey an individual's intentions and psychological states. To study facial muscle movements more comprehensively, Ekman and Friesen established the Facial Action Coding System (FACS)~\cite{FACS}, which describes a classification of facial Action Units (AU) framework based on observed activation of muscles or muscle groups to encode facial expressions. Facial AU detection is the basis for describing facial expressions, and it has received more and more attention due to its broad application prospects in sentiment analysis, human-computer interaction, etc~\cite{Imigue}.

Anatomical studies have indicated that a certain AU does not exist independently from other AUs and there are innumerable correlations between AUs~\cite{AUsurvey}. A certain expression of a person is the result of the synergy of many AUs. Therefore, learning the correlations between AUs is crucial for AU detection and there are several studies that explored the correlation information between AUs to improve the performance of AU detection. Earlier, most of the methods tried to design machine learning models based on Dynamic Bayesian Network (DBN)~\cite{YanTong2007FacialAU}, Restrict Boltzmann Machine (RBM)~\cite{ZihengWang2013CapturingGS,ShanWu2016MultipleFA}, etc., which rely on hand-crafted features. With the rapid development of deep learning, a large number of powerful correlation learning methods for AU detection have emerged. Representative works are based on Long-Short-Term-Memory (LSTM)~\cite{LP-Net}, message passing~\cite{DSIN,HMP-PS}, Transformer~\cite{FAUDT, FAN-Trans}, and Graph Convolutional Network (GCN)~\cite{SRERL,AU-GCN,UGN-B, ME-graphAU}. Compared to other methods, GCNs have demonstrated greater flexibility and a stronger ability to model complex correlations. Usually, existing methods~\cite{SRERL,AU-GCN,chang2022knowledge} fixed the correlations between AUs based on the summary of FACS~\cite{FACS} or statistical rules on the specific dataset (shown in Figure \ref{fig.1} (a)). However, these two ways of modeling the correlations are flawed. The AU correlations based on the FACS summary introduce subjective judgment of the researchers inevitably, which may ignore much correlation information between AUs. Due to the common class imbalance problem in the dataset, the AU correlations based on the statistical rules of the dataset tend to a biased result, which leads to a limited ability to model the actual correlations between AUs. In summary, the fixed graph structure leads to the inability to dynamically explore the potential correlations between AUs. There are alternative methods employ a fully connected graph to iteratively learn the correlations for each pair of AUs in a data-driven manner (shown in Figure \ref{fig.1} (b)), such as \cite{UGN-B,ME-graphAU}. Although this strategy alleviates the shortcomings of the fixed graph structure, modeling correlations for all AU pairs ergodically using fully connected graph structure is prone to becoming overly redundant and computationally expensive. Moreover, this will lead to a higher demand on the amount of data, which is not friendly to difficult-to-collect AU-related data. How to efficiently and effectively explore the potential correlation information between AUs is a challenging issue to be solved.

Besides, considering that AUs are active in diverse regions of the face, region learning is also a hotspot in AU detection. Various network structures have been designed to perform effective feature extraction on local regions of facial data~\cite{JPML,DRML,EAC-Net,PIAP-DF,EnriqueSnchezLozano2018JointAU}. Since different AUs correspond to varying scales, some researchers proposed to use multi-scale learning to improve the performance of AU detection~\cite{JAA-Net2018,JAA-Net2020,ARL,han2018optimizing}. However, to date, few studies have explored the role of multi-scale learning in correlation information learning, which represents a gap in research.

Considering the above limitations, we propose a multi-scale promoted self-adjusting correlation learning network for AU detection. Inspired by Occam's Razor principle, which states that the simplest explanation is often the best, we argue that a simpler and more streamlined graph is often preferable to one that is overly complex and difficult to navigate. By embracing simplicity and flexibility in our graph structures, we propose a novel self-adjusting AU-correlation learning (SACL) method, which aims to mine more potential AU correlation information with less computational effort by adaptively exploring the different graph structures at different stages. We suppose that two aspects should be considered in SACL: AU motion itself and AU emotion representation. Features of shallow layers contain rich fine-grained changes caused by muscle motions, and the similarity of motions can infer partial correlation information between AUs. With the deepening of layers, the features of nodes are gradually tend to be semantic, and the high-level semantic information encompasses a more abstract representation of emotions of each AU. In that way, emotion representation obtained from the deep layer can be used to supplement the implicit AU correlations that cannot be captured in muscle motions. As a result, richer AU correlation information can enhance the representation and help the model handle challenging situations of AU detection. By comparison, our method can adaptively capture more potential AU correlation information (such as AU1 and AU15) than the constructed graph based on the FACS summary or the statistical rules of the dataset (shown in Figure \ref{visualization}). Meantime, our method is less computationally expensive than calculating the correlation information for each pair of AUs. Specifically, our number of parameters and FLOPs are only 28.7\% and 12.0\% of the current optimal method \cite{ME-graphAU}, respectively (shown in Table \ref{tab:model complexity}), echoing the Ockham's Razor principle. The novelty of our SACL in comparison to existing AU correlation learning is illustrated in Figure \ref{fig.1}. In addition, we introduce a simple yet effective multi-scale feature learning (MSFL) method that concatenates the features extracted at each stage of the backbone to promote SACL. By integrating the correlation information with the multi-scale features, we can obtain a more powerful representation for final AU detection. 

The contributions of this paper can be summarized as follows:
\begin{itemize}
    \item This work propose a novel SACL method for AU detection, which can mine more implicit and precise AU correlation information with fewer parameters and FLOPs, so as to improve the performance of AU detection.
    \item We explore the role of multi-scale learning in correlation information learning, and design a simple yet effective MSFL method to promote SACL and obtain a more informative and robust representation of AUs for final AU detection.
    \item Extensive experiments show that the proposed method outperforms the state-of-the-art methods on the widely used AU detection benchmark datasets, namely, the BP4D~\cite{BP4D} and DISFA~\cite{DISFA}.
\end{itemize}

In the rest of the paper, Section \ref{sec:related} reviews related work focusing on AU correlation learning, region learning and using additional relevant data. Section \ref{sec:method} introduces the details of our method and the loss functions utilized in training. Section \ref{sec:experiment} provides comparisons of experimental results between our method and the state-of-the-art methods, demonstrating the superiority of our method, and also includes the details of ablation studies. Finally, conclusions and future work are given in Section~\ref{sec:conclusion}.

\section{Related work}
\label{sec:related}

\begin{table*}[t]
\centering
\caption{Summary of the representative AU detection methods in terms of AU correlation learning, region learning, using additional relevant data.} 
\label{tab:category}

{\begin{tabular} {c c c c} 
\toprule[1pt]
Category       &       Method   &   Venue   &   Details                    \\
\midrule
correlation learning    &    LP-Net~\cite{LP-Net}     &  CVPR 2019   &  using LSTM for local features      \\
without using graphs                                                   &  FAUDT~\cite{FAUDT}   &  CVPR 2021  &  using Transformer for AU-specific features       \\
~                                                   &    FAN-Trans~\cite{FAN-Trans}  &  WACV 2023  &  using Transformer with a drop mechanism for AU-specific features        \\
\cmidrule(lr){1-4}
~                                                   &   DSIN~\cite{DSIN}     &  ECCV 2018   &  using a message passing algorithm between AUs  \\
~                                                   &    SRERL~\cite{SRERL}  &  AAAI 2019   &  using a fixed graph based on the statistical rules       \\
correlation learning                                &    AU-GCN~\cite{AU-GCN}  &  MMM 2020  &  using a fixed graph based on the statistical rules        \\
using graphs                                        &    UGN-B~\cite{UGN-B}   &  AAAI 2021  &  using a fully connected graph with a probabilistic mask     \\
~                                                   &    HMP-PS~\cite{HMP-PS}    &  CVPR 2021  &  using a performance-driven hybrid message passing algorithm      \\
~                                                   &    ME-GraphAU~\cite{ME-graphAU}  &  IJCAI 2022  & using a fully connected graph with multi-dimensional edges        \\
\cmidrule(lr){1-4}
\multirow{6}{*}{region learning}           &    JPML~\cite{JPML}   &  CVPR 2015   &   selecting a sparse subset of face patches to exploit group sparsity      \\
~                                                   &    DRML~\cite{DRML}   &  CVPR 2016  &  using the region layer to focus on important face regions       \\
~                                                   &    EAC-Net~\cite{EAC-Net}  &  TPAMI 2018  &  using enhancing layers and cropping layers for AU-specific facial regions        \\
~                                                   &    ARL~\cite{ARL}     &  TAC 2019   &  using channel-wise and spatial attentions for learning AU region features   \\
~                                                   &    J{\^A}A-Net~\cite{JAA-Net2020}   &  IJCV 2021  &  using a hierarchical and multi-scale region layer for multi-scale learning      \\
~                                                   &    PIAP~\cite{PIAP-DF}  &  ICCV 2021  & using a pixel-wise interest learning to obtaining fine-grained features        \\
\cmidrule(lr){1-4}
using additional   &    SEV-Net~\cite{SEV-Net}  &  CVPR 2021  &  introducing the AU semantic text descriptions        \\
relevant data      &    KDSRL~\cite{chang2022knowledge}  &  CVPR 2022       &  using a larger dataset BP4D+ for pre-training  \\
~     &    AUFM~\cite{AUFM}  &  CVPR 2023 &  introducing the biomechanical guidance based on 3D surface mesh       \\

\bottomrule[1pt]
\end{tabular}}
\end{table*}

Due to the rapid development of computer vision in recent years, facial AU detection has attracted more and more researchers, and several effective methods have been proposed. Firstly, we focus on the related AU correlation learning. Then, we give an overview of the literature related to region learning. Finally, we present some recent competitive methods that use additional relevant data. Table \ref{tab:category} shows a rough summary and classification of representative methods.

\subsection{AU correlation learning}

According to the analysis of various facial expressions, there are symbiosis or mutual exclusion correlations between AUs. Therefore, AU correlation learning is very necessary. In the early works~\cite{YanTong2007FacialAU,ZihengWang2013CapturingGS,ShanWu2016MultipleFA}, researchers explored the correlations between AUs using DBN or RBM. However, these works are limited because they are based on hand-crafted features. Niu \etal~\cite{LP-Net} used LSTM for correlation learning of local features generated by ResNet~\cite{ResNet}. Jacob and Stenger~\cite{FAUDT} introduced Transformer to conduct correlation learning among features extracted from branches specific to each AU. Yang \etal~\cite{FAN-Trans} proposed a learnable attention drop mechanism in the Transformer block to learn the correlation between different AUs. Corneanu \etal~\cite{DSIN} used a message passing algorithm similar to a graphical model inference approach in later stages to capture the correlations between AUs. Song \etal~\cite{HMP-PS} proposed a performance-driven hybrid message passing algorithm that combines different types of messages to mine more possible AU correlation information. In addition to the above, there are several GCN-based methods. Li \etal~\cite{SRERL} embedded the semantic relationship of AUs into the deep neural network to enhance the feature representation. The features of AUs are taken as nodes to construct a structured knowledge graph based on statistical rules in the datasets, and a GGNN~\cite{GGNN} is applied to transfer the information between nodes. Liu \etal~\cite{AU-GCN} obtained region of interest (ROI) features according to landmarks, which are also taken as nodes to construct a correlation graph based on the statistical characteristics. Then, the GCN is used for feature extraction. Song \etal~\cite{UGN-B} proposed an uncertain graph-based network that used a probabilistic mask applied to a fully connected graph to select useful edges and suppress noisy edges, thereby capturing the dependencies among AUs. Recently, Luo \etal~\cite{ME-graphAU} learned multi-dimensional edge features based on an ergodic fully connected graph to describe the correlations between each pair of AUs using GGCN~\cite{GGCN}.

\subsection{Region learning}

In addition, AU is known to be active in diverse regions of the face. Focusing on these AU-related regions is also helpful for AU detection. Zhao \etal~\cite{JPML} explored patches centered at facial landmarks and selected a sparse subset of face patches to exploit group sparsity. Zhao \etal~\cite{DRML} proposed a region layer, which induces important facial regions by uniformly slicing the feature map from shallow layers into patches and using independent convolution kernels on each patch, forcing the learned weights to capture the structural information of the face. Li \etal~\cite{EAC-Net} introduced enhancing layers and cropping layers into a pretrained CNN to perform deeper feature learning for AU-specific facial regions. Besides, the size of ROIs corresponding to each AU is varied, which is not taken into account when only extracting features from regions of fixed size. Therefore, multi-scale learning is introduced to extract richer features. Shao \etal~\cite{JAA-Net2018, JAA-Net2020} improved the region layer in~\cite{DRML} and proposed a hierarchical and multi-scale region learning layer. The novel region learning layer is composed of three region layers in~\cite{DRML}, and each layer uniformly sliced the feature maps into a different number of patches. The feature maps extracted from the three region layers are concatenated to obtain the hierarchical and multi-scale features. Shao \etal~\cite{ARL} adopted the hierarchical and multi-scale features proposed in \cite{JAA-Net2018, JAA-Net2020} and introduced channel-wise and spatial attentions to adaptively learn region features related to AUs. Recently, Tang \etal~\cite{PIAP-DF} proposed a pixel-wise interest learning method with pixel-level attention for each AU to extract fine-grained features.

\begin{figure*}[t]
  \centering
  \includegraphics[width=1\linewidth]{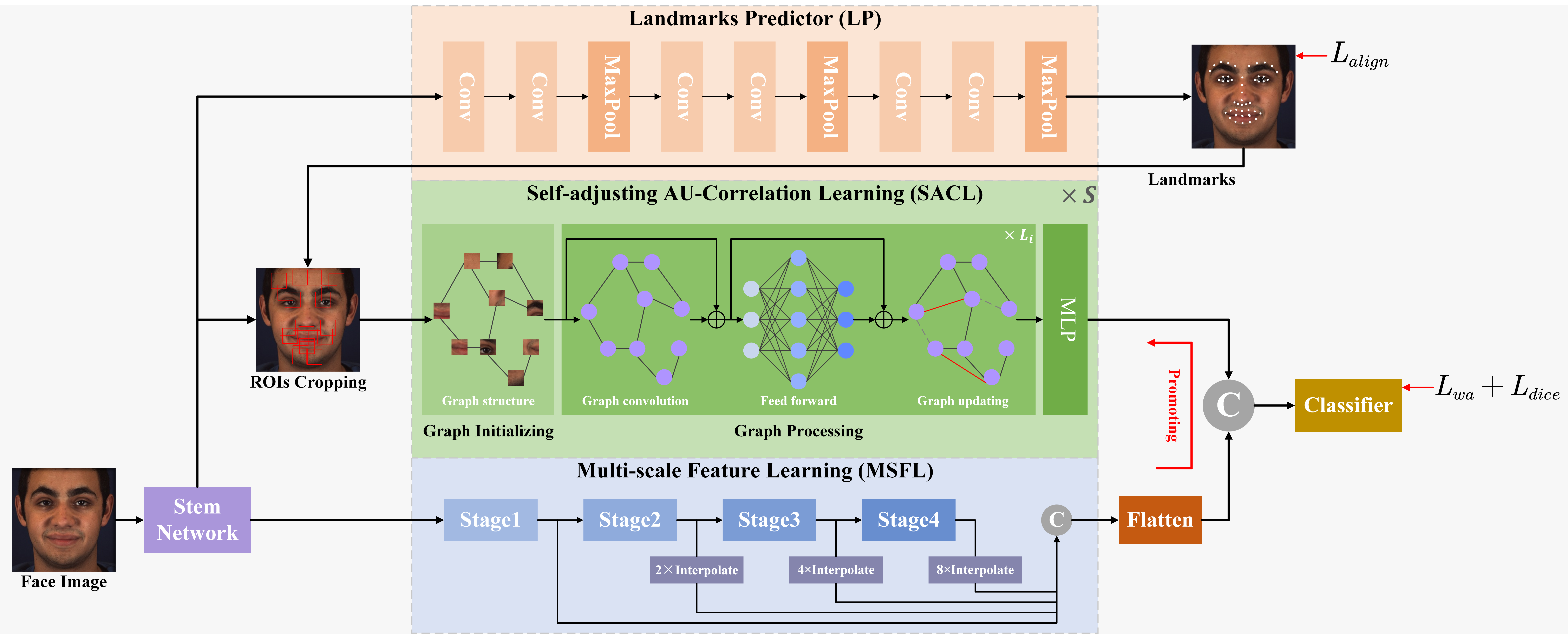}
  \caption{An overview of our method for AU detection, which mainly consists of stem network, MSFL, LP, and SACL. The face image is firstly fed into the stem network to obtain basic features. Then, the basic features are further processed through the remaining three modules. Finally, the multi-scale features and AU correlation information are integrated for the final AU detection. `C' denotes concatenation operation. }
  \label{pipeline}
\end{figure*}


\subsection{Using additional relevant data}
In recent years, there are also some methods to introduce additional relevant data to improve the performance of AU detection. Yang \etal~\cite{SEV-Net} summarized the semantically rich text descriptions of AUs in FACS~\cite{FACS} and used text description information to assist in the generation of attention maps to capture AU-specific regional features. Chang \etal~\cite{chang2022knowledge} built an AU correlation graph based on the summary of FACS~\cite{FACS} and relied on this prior knowledge to perform self-supervised representation learning, where the larger BP4D+~\cite{BP4D+} dataset was used for pre-training. Cui \etal~\cite{AUFM} constructed a 3D physical branch, introducing biomechanical guidance based on 3D surface mesh to enhance the performance of AU detection.

In contrast to these existing methods, our proposed SACL adaptively adjusts the AU correlation graph as the network deepens, effectively leveraging the characteristics of different levels of AU motion and emotion representation information extracted in different stages of the network. SACL captures more accurate AU correlation information than fixed graphs and saves computational effort compared to data-driven fully connected graphs. Additionally, we introduce MSFL to further improve SACL performance, which is rarely explored by existing methods. By integrating the features output from SACL and MSFL, we can obtain a more robust representation for final AU detection. It is worth noting that our method does not use additional relevant data during the training process.

\section{Methodology}
\label{sec:method}

\subsection{Overview}
The architecture of the proposed method is shown in Figure \ref{pipeline}. It is mainly composed of four modules: stem network, MSFL, landmarks predictor (LP), and SACL. Firstly, the face image is input to the stem network for extracting the basic features. Then, the basic features are fed into the following three modules. Since different AUs have various sizes, MSFL is designed to obtain multi-scale global features of faces and explore its role in AU correlation learning. LP is used to predict the locations of facial landmarks, which are further utilized to calculate AU-specific centers. According to the centers of AUs, the corresponding ROIs are cut out from the basic features extracted by stem network, which are fed into SACL. SACL is designed to capture correlation information between AUs, which updates the graph structure after each feature mapping to mine more implicit correlation information. Finally, the features from MSFL and SACL are integrated to realize the final AU detection.
\begin{table*}[t]
\begin{center}
\caption{The definitions of the AU center locations and the formula for calculating them through Landmark coordinates. `Scale’ represents the distance between the two inner corners of the eyes.}

\label{tab:aucenter}
\centering
\begin{tabular}{c c c c c}

\toprule[1pt]

AU index & AU description & AU center location & Landmark index & AU center formula\\

\midrule
 
1   & Inner brow raiser     & 1/2 scale above inner brow    & 4, 5      & (x, y-scale/2)    \\
2   & Outer brow raiser     & 1/3 scale above outer brow    & 1, 8      & (x, y-scale/3)    \\
4   & Brow lowerer          & 1/3 scale below brow center   & 2, 7      & (x, y+scale/3)    \\
6   & Cheek raiser          & 1 scale below eye bottom      & 24, 29    & (x, y+scale)      \\
7   & Lid tightener         & Eye                           & 21, 26    & (x, y)        \\
9   & Nose wrinkler         & 1/2 scale above nose bottom   & 15, 17    & (x, y-scale/2)    \\
10  & Upper lip raiser      & Upper lip center              & 43, 45    & (x, y)        \\
12  & Lip corner puller     & Lip corner                    & 31, 37    & (x, y)        \\
14  & Dimpler               & Lip corner                    & 31, 37    & (x, y)        \\
15  & Lip corner depressor  & Lip corner                    & 31, 37    & (x, y)        \\
17  & Chin raiser           & 1/2 scale below lip           & 39, 41    & (x, y+scale/2)    \\
23  & Lip tightener         & Lip center                    & 34, 40    & (x, y)        \\
24  & Lip pressor           & Lip center                    & 34, 40    & (x, y)        \\
25  & Lips part             & Lip center                    & 34, 40    & (x, y)        \\
26  & Jaw drop              & 1/2 scale below lip           & 39, 41    & (x, y+scale/2)    \\

\bottomrule[1pt]
\end{tabular}
\end{center}
\end{table*}

\subsection{Stem Network and Multi-scale Feature Learning}


Our method adopts a flexible framework design pattern so that a variety of different pyramid Transformer models~\cite{Swin,ResT,MPViT,Crossformer,PVT} can be used as the backbone for building the stem network and MSFL to extract global information of the entire face. In general, the overall architecture of the common pyramid Transformer is mainly composed of a stem network and four stages that output features with various receptive fields, as shown at the bottom of Figure \ref{pipeline}. 

To extract basic features from the input face image with size of $H \times W \times 3$, the stem network is simply consistent with the selected pyramid Transformer model consisting of several convolutional layers with overlapping. Then, these obtained basic features $F \in \mathbb{R}^{\frac{H}{4} \times \frac{W}{4} \times d_0}$ will be fed into subsequent modules for further processing. Considering that different AUs correspond to various region sizes, MSFL is introduced into the four stages of pyramid Transformer to extract more informative features and enrich the receptive field of features. Specifically, the features extracted by the four stages of the pyramid Transformer can be represented as $F_1 \in \mathbb{R}^{\frac{H}{4} \times \frac{W}{4} \times d_0}$, $F_2 \in \mathbb{R}^{\frac{H}{8} \times \frac{W}{8} \times 2d_0}$, $F_3 \in \mathbb{R}^{\frac{H}{16} \times \frac{W}{16} \times 4d_0}$, and $F_4 \in \mathbb{R}^{\frac{H}{32} \times \frac{W}{32} \times 8d_0}$, respectively. The features extracted by the last three stages are interpolated to align the spatial size with the features extracted by the first stage, and then, these aligned features are concatenated to form multi-scale features $A \in \mathbb{R}^{\frac{H}{4} \times \frac{W}{4} \times D}$, where $D = d_0 + 2d_0 + 4d_0+ 8d_0$. This process is depicted in the blue area in Figure \ref{pipeline} and can be expressed as 
\begin{equation}
    A = Concat (F_1, 2 \times F_2, 4 \times F_3, 8 \times F_4),
\end{equation}
Where, $2 \times$, $4 \times$, and $8 \times$ represent two-fold, four-fold, and eight-fold interpolation in the spatial dimension, respectively.

\subsection{Landmarks Predictor}
In order to locate the AU-specific centers for cropping ROIs, Landmarks Predictor is designed to extract the shape information of the face and predict the locations of the facial landmarks. LP contains three consecutive blocks, each consisting of two convolutional layers and a max-pooling layer (shown in the orange area of Figure \ref{pipeline}). After each convolution operation, the features are processed with Batch Normalization~\cite{BN} and non-linearly transformed by the ReLU activation function~\cite{RELU}. The features obtained from the last max-pooling layer are flattened and fed into the fully connected layer (omitted in Figure \ref{pipeline}) for landmarks prediction. The output dimension of the fully connected layer is $N_{land}$, which is the number of landmark coordinates.

\subsection{Self-adjusting AU-correlation Learning}
\begin{figure}[t]
    \centering
    \includegraphics[width=1\linewidth]{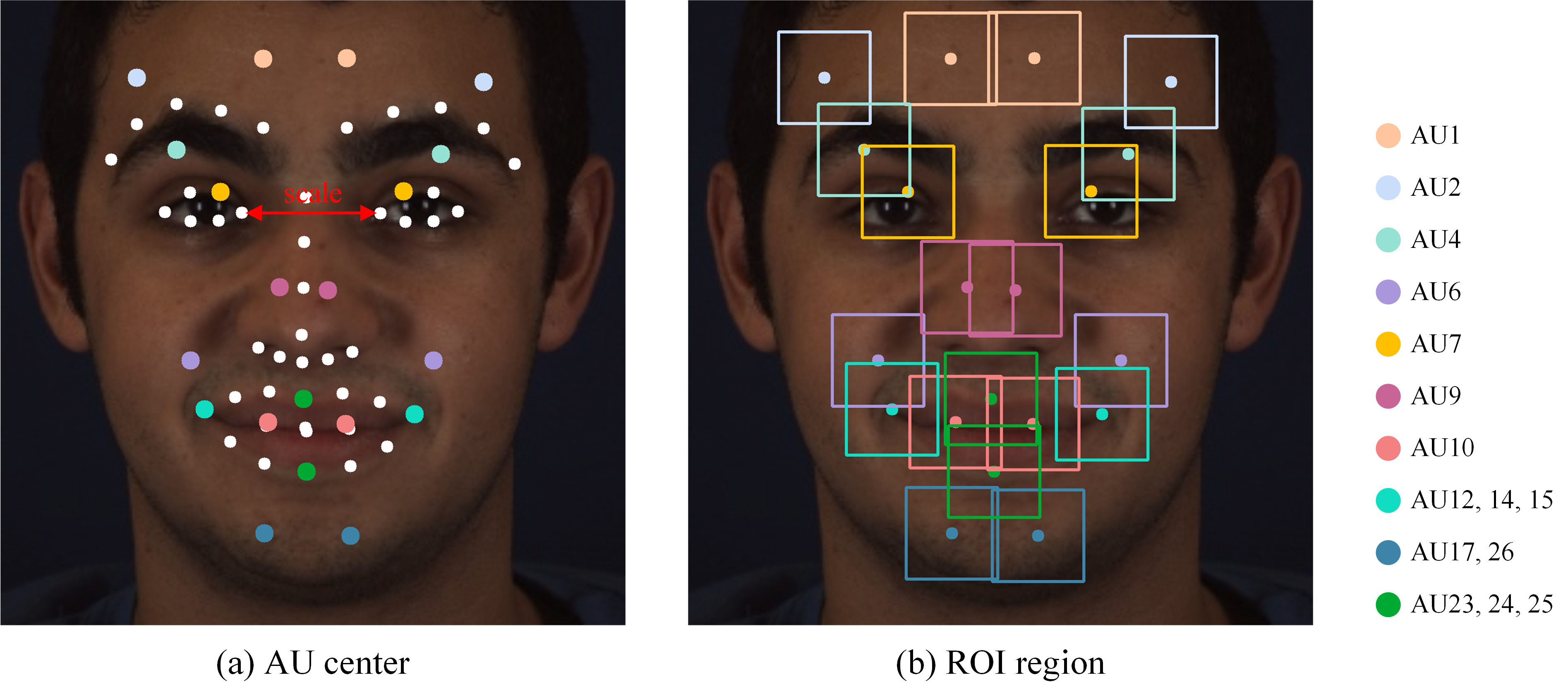}
    \caption{
    (a) The AU centers calculated based on the landmark coordinates. The white dots are landmarks. The red arrow corresponds to the `scale' in Table \ref{tab:aucenter} and donates the distance between the two inner corners of the eyes. (b) The ROI cropping region obtained based on the AU centers and $\xi$. 
    }
    \label{AUcenter}
\end{figure}

Referring to \cite{JAA-Net2018,JAA-Net2020}, based on the predicted landmark coordinates, the AU-specific centers are determined. The definition rules and calculation formula of each AU-specific center location are shown in Table \ref{tab:aucenter}, and the visualization of AU-specific center location is shown in Figure \ref{AUcenter} (a). Because the spatial size of features obtained by the stem network differs from the input image, the coordinates of the landmarks need to be multiplied by a factor $\eta$ for scaling. Then, the ROIs are cropped from the basic features according to the centers of AUs, where the spatial size ratio of the ROIs to the basic features is $\xi$. The size of each ROI is $(\xi \times \frac{H}{4}) \times (\xi \times \frac{W}{4}) \times d_0$, where $d_0$ is the number of basic feature channels. Figure \ref{AUcenter} (b) visualizes this operation. Here, the basic feature maps are replaced with the input original face image to get a more intuitive expression.

The proposed SACL is illustrated in the green area of Figure \ref{pipeline}. Firstly, the features of obtained ROIs are flattened, and each ROI is taken as a graph node $v_i \in \mathbb{R}^{d_1}$, where $d_1 = (\xi \times \frac{H}{4}) \times (\xi \times \frac{W}{4}) \times d_0$. Thus, the node set can be represented as $V = \{v_1, v_2, \cdots, v_{N_{ROI}}\}$, where $N_{ROI}$ is the number of ROIs. In the feature domain, the correlations between nodes are negatively correlated with the feature distance, that is, the smaller the feature distance, the greater the correlation, and vice versa. Based on the above characteristics, the $K$-nearest neighbors (via the KNN algorithm) are calculated for each node to get the edge set $E$. Finally, the graph of AUs $G = (V, E)$ can be constructed.

Given a graph structure data $G \in \mathbb{R}^{N_{ROI} \times d_1}$, the graph convolution can be utilized to aggregate and communicate information between adjacent nodes for learning the correlation information between AUs. In order to increase the diversity of features, fully connected layers are added before and after graph convolution, and the activation function for nonlinear mapping is applied. In addition, the residual connection is introduced to avoid a vanishing gradient. Specifically, this process can be expressed as
\begin{equation}
    G^{\prime} = \sigma(GCN(GW_{before}))W_{after} + G,
\end{equation} 
where $G^{\prime} \in \mathbb{R}^{N_{ROI} \times d_1}$, $\sigma$ is the GeLU~\cite{GeLU} activation function, $W_{before}$ and $W_{after}$ are the learnable weights of the fully connected layers before and after graph convolution, respectively.

Inspired by \cite{ViG}, a feed forward network with two fully connected layers is introduced to alleviate the over smoothing phenomenon in the graph convolution. We also introduce residual connections. Specifically, this process can be expressed as
\begin{equation}
    G^{\prime\prime} = \sigma(G^{\prime}W_1)W_2 + G^{\prime},
\end{equation} 
where $G^{\prime\prime} \in \mathbb{R}^{N_{ROI} \times d_1}$, $W_1$ and $W_2$ are the learnable weights of the fully connected layers. After each feed forward network, the $K$-nearest neighbors are recalculated using the obtained features to update the graph structure, and thereby, self-adjusting learning can be realized. After several graph processing, a fully connected layer is introduced for feature dimension expansion, and then the graph is re-initialized. 

It can be seen from Figure \ref{pipeline} that SACL consists of $S$ stages, each of which includes initializing of graph structure, $L_i$ graph processing blocks and a fully connected layer. The final features of the $S_{th}$ stage is $B \in \mathbb{R}^{N_{ROI} \times D}$. By constantly updating the graph structure, SACL can make use of the similarities and differences between shallow layer muscle motion information and deep layer emotion representation information to construct different graph structures. In that way, the information between related AUs can be sufficiently communicated and aggregated, and more effective and implicit correlation information can be extracted.

\subsection{AU Detection}
Considering the inconsistent sizes of the multi-scale features $A$ and the AU correlation features $B$, a transformation is performed on $A$ to integrate the features of two branches. Since the tokens in the Transformer features are treated as nodes, $A$ is flattened in the spatial dimension to get $A^{\prime} \in \mathbb{R}^{N_{MS}\times D}$, where $N_{MS}=\frac{H}{4} \times \frac{W}{4}$. Finally, as illustrated in Figure \ref{pipeline}, $A^{\prime}$ and $B$ are concatenated together to obtain $C \in \mathbb{R}^{(N_{MS}+N_{ROI})\times D}$ with comprehensive effective information, which is finally fed into the classifier with fully connected layers for the final AU detection.

\subsection{Loss Function}
AU detection is regarded as a multi-label binary classification problem, and many previous methods~\cite{JAA-Net2018,JAA-Net2020,LP-Net,AU-GCN,FAUDT} adopted a weighted multi-label cross-entropy loss function for supervision. However, the difficulty level of detecting various AUs is different, but this loss function did not fully take into account these differences among AUs. For example, a well-designed loss function could push the network to learn the features of AUs which are hard to be identified correctly. In this paper, the weighted asymmetric loss function \cite{ME-graphAU} is employed to focus on activate AUs and inactivate AUs that are hard to be correctly recognized. It can be formulated as
\begin{equation}
    \mathcal{L}_{wa} = - \frac{1}{N_{AU}} {\sum_{i=1}^{N_{AU}} \omega_{i}[y_i \log(p_i) + (1-y_i) p_i \log(1-p_i)]},
\end{equation}
where $y_i$ is the ground truth of the $i_{th}$ AU, $p_i$ is the corresponding prediction probability, and $N_{AU}$ is the number of AUs. $\omega_i$ is the weight for alleviating the imbalance problem in the dataset, and it can be formulated as $\omega_i = N_{AU} (1/r_i) / {\sum_{j=1}^{N_{AU}} (1/r_j)}$, where $r_i$ is the occurrence rate of the $i_{th}$ AU.

Moreover, considering that AU detection strongly biases towards non-occurrence, a weighted multi-label dice loss \cite{JAA-Net2018, JAA-Net2020} is introduced. It can be formulated as
\begin{equation}
    \mathcal{L}_{dice} = \frac{1}{N_{AU}} {\sum_{i=1}^{N_{AU}} \omega_i (1 - \frac{2 y_i p_i + \varepsilon}{y_i^2 + p_i^2 + \varepsilon})},
\end{equation}
where $\varepsilon$ is a smooth term.

Additionally, a face alignment loss is utilized to supervise the LP. It can be formulated as
\begin{equation}
    \mathcal{L}_{land} = \frac{1}{2d_{o}^2} {\sum_{i=1}^{N_{land}} [(c_{2i-1}-\hat{c}_{2i-1})^2 + (c_{2i}-\hat{c}_{2i})^2]},
\end{equation}
where $c_{2i-1}$ and $c_{2i}$ are the ground truth of x coordinate and y coordinate of the $i_{th}$ landmark, respectively, and $\hat{c}_{2i-1}$ and $\hat{c}_{2i}$ are the corresponding predictions. $d_o$ is the inter-ocular distance of ground truth for normalization~\cite{ZhiwenShao2016LearningDR,JAA-Net2018,JAA-Net2020,ZhiwenShao2020DeepML,liu20193d,liu20203d}.

Combined with the above loss functions, our overall loss function can be formulated as
\begin{equation}
    \mathcal{L} = \lambda_1 \mathcal{L}_{wa}+ \lambda_2 \mathcal{L}_{dice}+ \lambda_3 \mathcal{L}_{land},
\end{equation}
where $\lambda_1$, $\lambda_2$ and $\lambda_3$ are the trade-off parameters.

\section{Experimental Results}
\label{sec:experiment}
In this section, we provide evaluation results on several datasets widely used for AU detection and present a detailed analysis of the experimental results.


\subsection{Settings}
\subsubsection{Dataset}
We evaluate the proposed method on two benchmark datasets, namely, the BP4D~\cite{BP4D} and DISFA~\cite{DISFA}. 
\begin{itemize}
\item \textbf{BP4D} contains 2D and 3D data of spontaneous facial expressions from 41 subjects (23 females and 18 males). Each subject was given well-validated emotion induction through 8 different tasks. There are totally 328 videos, and about 140,000 frames are annotated with AU labels. Each frame is also annotated with 49 landmarks detected by SDM~\cite{SDM}. 

\item \textbf{DISFA} records facial expression data from 27 subjects (12 females and 15 males) while watching video clips intended to elicit spontaneous emotion expression. About 130,000 frames are available in DISFA, and each frame is annotated with AU intensity from 0 to 5. According to \cite{DRML}, if the AU intensity is equal or greater than 2, it is considered to be present; otherwise, it is not present. Each frame in DISFA is also annotated with 66 landmarks detected by AAM~\cite{AAM}, and we map them to 49 landmarks for consistency with BP4D~\cite{JAA-Net2020}.
\end{itemize}

Following the experiment setting of \cite{DRML,JAA-Net2018,JAA-Net2020, ME-graphAU}, a subject-exclusive 3-fold cross-validation is performed on BP4D, and the best-trained model on BP4D is then fine-tuned on DISFA. The partition details of each fold of BP4D and DISFA are shown in Table \ref{tab:3fold}.

\begin{table}[t]
\begin{center}
\caption{The partition details of BP4D~\cite{BP4D} and DISFA~\cite{DISFA} in the subject-exclusive 3-fold cross-validation, where two folds are used for training and the remaining fold is used for testing.}

\label{tab:3fold}
\centering
\begin{tabular}{p{0.8cm} p{6.7cm}}

\toprule[1pt]
\multicolumn{2}{c}{BP4D}    \\
\cmidrule(lr){1-2}
\multirow{2}{*}{Fold-1}& F001, F002, F008, F009, F010, F018, F016, F023, \\
                     ~ & M001, M004, M007, M008, M012, M014.    \\ 
\cmidrule(lr){1-2}
\multirow{2}{*}{Fold-2}& F003, F005, F011, F013, F020, F022,  \\
                     ~ & M002, M005, M010, M011, M013, M016, M017, M018.  \\ 
\cmidrule(lr){1-2}
\multirow{2}{*}{Fold-3}& F004, F006, F007, F012, F014, F015, F017, F019, F021,  \\
                     ~ & M003, M006, M009, M015.  \\

\midrule[0.6pt]

\multicolumn{2}{c}{DISFA}    \\
\cmidrule(lr){1-2}
\multirow{2}{*}{Fold-1}& SN001, SN002, SN009, SN010, SN016, SN026, SN027, \\
                     ~ & SN030, SN032. \\
\cmidrule(lr){1-2}
\multirow{2}{*}{Fold-2}& SN006, SN011, SN012, SN013, SN018, SN021, SN024, \\
                     ~ & SN028, SN031.  \\
\cmidrule(lr){1-2}
\multirow{2}{*}{Fold-3}& SN003, SN004, SN005, SN007, SN008, SN017, SN023, \\
                     ~ & SN025, SN029. \\

\bottomrule[1pt]
\end{tabular}
\end{center}
\end{table}

\subsubsection{Data pre-processing} For each original face image, we conduct similarity transformation using landmarks. This transformation includes in-plane rotation, uniform scaling, and translation, which preserves the shape and expression information while reducing the variations of pose and scale. Then, the aligned faces are resized to 256 $\times$ 256. To increase the diversity of the data, we randomly crop these face images to 224 $\times$ 224 as the inputs. Furthermore, random horizontal flipping and random color jittering (contrast, brightness) for data augmentation are also introduced.

\subsubsection{Implementation details} We choose the ResT-Lite~\cite{ResT} (pretrained on ImageNet-1k~\cite{ImageNet}) as the backbone (for our stem network and MSFL) due to its lightweight. Also, it has a similar structure with ResNet~\cite{ResNet}, which has been widely used in previous AU detection methods~\cite{HMP-PS,LP-Net,SRERL,ME-graphAU}. During the training, we set batch size to 16 and employ an SGD optimizer~\cite{SGD} with a Nesterov momentum~\cite{Nesterov} of 0.9 and a weight decay of 0.0005. We use a cosine decay learning rate scheduler with a maximum learning rate of 0.001 for 12 epochs, and the first epoch is used for linear warm-up. Besides, We use gradient clipping~\cite{Gradient} with a max norm of 5. Table \ref{tab:parameter settings} shows the parameter settings of our model. The parameters about the loss function are set as $\lambda_1=1$, $\lambda_2=1$, $\lambda_3=0.5$. All our experiments are conducted using an Nvidia RTX 3090 GPU based on the open-source PyTorch~\cite{Pytorch} platform.


\begin{table}[t]
\centering
\caption{The parameter settings about the model.} 
\label{tab:parameter settings}


{\begin{tabular} { c  c } 
\toprule[1pt]
Parameter       &       Setting                     \\
\midrule
$S$             &       4                           \\
$L_i$           &       2, 2, 6, 2                  \\
$\eta$          &       0.25                        \\
$\xi$           &       0.14                        \\
$D$             &       960                         \\
$d_0$           &       64                          \\
$K$-nearest     &       9                           \\
$N_{land}$      &       49                          \\
$N_{ROI}$       &       18 (BP4D) / 16 (DISFA)      \\
$N_{AU}$        &       12 (BP4D) / 8 (DISFA)       \\
GCN type        &       Max-Relative GraphConv~\cite{Mr}      \\

\bottomrule[1pt]
\end{tabular}}
\end{table}



\begin{table*}[t]
\begin{center}
\caption{F1-score and accuracy results for 12 AUs on BP4D\cite{BP4D}. The best results for each column are bolded. \% is omitted.}

\label{tab:BP4D result}
\centering
\begin{tabular}{p{1.2cm} p{2.7cm} p{0.5cm} p{0.5cm} p{0.5cm} p{0.5cm} p{0.5cm} p{0.5cm} p{0.5cm} p{0.5cm} p{0.5cm} p{0.5cm} p{0.5cm} p{0.5cm} p{0.55cm}}

\toprule[1pt]
\multirow{2}{*}{Metrics} &
\multirow{2}{*}{Method}&
\multicolumn{12}{c}{AU} & \multirow{2}{*}{\textbf{Avg.}}\\
  
\cmidrule(lr){3-14}

~& ~& \multicolumn{1}{c}{1} & \multicolumn{1}{c}{2} & \multicolumn{1}{c}{4} & \multicolumn{1}{c}{6} & \multicolumn{1}{c}{7} & \multicolumn{1}{c}{10} & \multicolumn{1}{c}{12} & \multicolumn{1}{c}{14} & \multicolumn{1}{c}{15} & \multicolumn{1}{c}{17} & \multicolumn{1}{c}{23} & \multicolumn{1}{c}{24} & ~ \\
\midrule
 
\multirow{17}{*}{F1 score} & JPML~\cite{JPML} & 32.6 & 25.6 & 37.4 & 42.3 & 50.5 & 72.2 & 74.1 & 65.7 & 38.1 & 40.0 & 30.4 & 42.3 & 45.9 \\

~ & DRML~\cite{DRML} & 36.4 & 41.8 & 43.0 & 55.0 & 67.0 & 66.3 & 65.8 & 54.1 & 33.2 & 48.0 & 31.7 & 30.0 & 48.3 \\

~ & EAC-Net~\cite{EAC-Net} & 39.0 & 35.2 & 48.6 & 76.1 & 72.9 & 81.9 & 86.2 & 58.8 & 37.5 & 59.1 & 35.9 & 35.8 & 55.9 \\ 

~ & ARL~\cite{ARL}  & 45.8 & 39.8 & 55.1 & 75.7 & 77.2 & 82.3 & 86.6 & 58.8 & 47.6 & 62.1 & 47.4 & 55.4 & 61.1 \\

~ & J{\^A}A-Net~\cite{JAA-Net2020} & 53.8 & 47.8 & 58.2 & 78.5 & 75.8 & 82.7 & 88.2 & 63.7 & 43.3 & 61.8 & 45.6 & 49.9 & 62.4 \\

~ & PIAP~\cite{PIAP-DF} & 54.2 & 47.1 & 54.0 & 79.0 & 78.2 & \textbf{86.3} & \textbf{89.5} & 66.1 & 49.7 & 63.2 & 49.9 & 52.0 & 64.1 \\
\cmidrule(lr){2-15}
~ & AUFM~\cite{AUFM} & 57.4 & \textbf{52.6} & \textbf{64.6} & 79.3 & \textbf{81.5} & 82.7 & 85.6 & 67.9 & 47.3 & 58.0 & 47.0 & 44.9 & 64.1 \\

~ & KDSRL~\cite{chang2022knowledge} & 53.3 & 47.4 & 56.2 & 79.4 & 80.7 & 85.1 & 89.0 & 67.4 & 55.9 & 61.9 & 48.5 & 49.0 & 64.5 \\
\cmidrule(lr){2-15}
~ & LP-Net~\cite{LP-Net} & 43.4 & 38.0 & 54.2 & 77.1 & 76.7 & 83.8 & 87.2 & 63.3 & 45.3 & 60.5 & 48.1 & 54.2 & 61.0 \\ 

~ & FAUDT~\cite{FAUDT} & 51.7 & 49.3 & 61.0 & 77.8 & 79.5 & 82.9 & 86.3 & 67.6 & 51.9 & 63.0 & 43.7 & \textbf{56.3} & 64.2 \\ 

~ & FAN-Trans~\cite{FAN-Trans} & 55.4 & 46.0 & 59.8 & 78.7 & 77.7 & 82.7 & 88.6 & 64.7 & 51.4 & \textbf{65.7} & \textbf{50.9} & 56.0 & 64.8 \\

\cmidrule(lr){2-15}

~ & AU-GCN~\cite{AU-GCN} & 46.8 & 38.5 & 60.1 & \textbf{80.1} & 79.5 & 84.8 & 88.0 & 67.3 & 52.0 & 63.2 & 40.9 & 52.8 & 62.8 \\

~ & SRERL~\cite{SRERL} & 46.9 & 45.3 & 55.6 & 77.1 & 78.4 & 83.5 & 87.6 & 63.9 & 52.2 & 63.9 & 47.1 & 53.3 & 62.9 \\ 

~ & UGN-B~\cite{UGN-B}  & 54.2 & 46.4 & 56.8 & 76.2 & 76.7 & 82.4 & 86.1 & 64.7 & 51.2 & 63.1 & 48.5 & 53.6 & 63.3 \\

~ & HMP-PS~\cite{HMP-PS} & 53.1 & 46.1 & 56.0 & 76.5 & 76.9 & 82.1 & 86.4 & 64.8 & 51.5 & 63.0 & 49.9 & 54.5 & 63.4 \\

~ & ME-GraphAU~\cite{ME-graphAU} & 52.7 & 44.3 & 60.9 & 79.9 & 80.1 & 85.3 & 89.2 & \textbf{69.4} & 55.4 & 64.4 & 49.8 & 55.1 & 65.5 \\ 
\cmidrule(lr){2-15}
~ & \textbf{Ours} & \textbf{57.8} & 48.8 & 59.4 & 79.1 & 78.8 & 84.0 & 88.2 & 65.2 & \textbf{56.1} & 63.8 & 50.8 & 55.2 & \textbf{65.6} \\
\midrule

\multirow{5}{*}{Accuracy}  & 

EAC-Net~\cite{EAC-Net} & 68.9 & 73.9 & 78.1 & 78.5 & 69.0 & 77.6 & 84.6 & 60.6 & 78.1 & 70.6 & 81.0 & 82.4 & 75.2 \\

~ & ARL~\cite{ARL}  & 73.9 & 76.7 & 80.9 & 78.2 & 74.4 & 79.1 & 85.5 & 62.8 & 84.7 & \textbf{74.1} & 82.9 & 85.7 & 78.2 \\

~ & UGN-B~\cite{UGN-B}  & 78.6 & 80.2 & 80.0 & 76.6 & 72.3 & 77.8 & 84.2 & 63.8 & 84.0 & 72.8 & 82.8 & \textbf{86.4} & 78.2 \\

~ & J{\^A}A-Net~\cite{JAA-Net2020}  & 75.2 & 80.2 & 82.9 & 79.8 & 72.3 & 78.2 & 86.6 & 65.1 & 81.0 & 72.8 & 82.9 & 86.3 & 78.6 \\

\cmidrule(lr){2-15}

~ & \textbf{Ours} & \textbf{81.4} & \textbf{83.6} & \textbf{84.1} & \textbf{80.5} & \textbf{77.4} & \textbf{80.9} & \textbf{86.7} & \textbf{66.2} & \textbf{85.2} & 74.0 & \textbf{84.6} & 84.6 & \textbf{80.8} \\

\bottomrule[1pt]
\end{tabular}
\end{center}
\end{table*}
\begin{table*}[t] 
\begin{center}
\caption{F1-score and accuracy results for 8 AUs on DISFA\cite{DISFA}. The best results for each column are bolded. \% is omitted.}

\label{tab:DISFA result}
\centering
\begin{tabular}{p{1.2cm} p{2.7cm} p{0.6cm} p{0.6cm} p{0.6cm} p{0.6cm} p{0.6cm} p{0.6cm} p{0.6cm} p{0.6cm} p{0.65cm}}

\toprule[1pt]
 \multirow{2}{*}{Metrics} &
 \multirow{2}{*}{Method}&
 \multicolumn{8}{c}{AU} & \multirow{2}{*}{\textbf{Avg.}}\\
  
\cmidrule(lr){3-10}

~& ~& \multicolumn{1}{c}{1} & \multicolumn{1}{c}{2} & \multicolumn{1}{c}{4} & \multicolumn{1}{c}{6} & \multicolumn{1}{c}{9} & \multicolumn{1}{c}{12} & \multicolumn{1}{c}{25} & \multicolumn{1}{c}{26} &  ~ \\
 \midrule
 
 \multirow{16}{*}{F1 score} & DRML~\cite{DRML} & 17.3 & 17.7 & 37.4 & 29.0 & 10.7 & 37.7 & 38.5 & 20.1 & 26.7 \\ 

~ & EAC-Net~\cite{EAC-Net} & 41.5 & 26.4 & 66.4 & 50.7 & \textbf{80.5} & \textbf{89.3} & 88.9 & 15.6 & 48.5 \\

~ & ARL~\cite{ARL} & 43.9 & 42.1 & 63.6 & 41.8 & 40.0 & 76.2 & 95.2 & 66.8 & 58.7 \\

~ & J{\^A}A-Net~\cite{JAA-Net2020} & \textbf{62.4} & 60.7 & 67.1 & 41.1 & 45.1 & 73.5 & 90.9 & 67.4 & 63.5 \\

~ & PIAP~\cite{PIAP-DF} & 50.2 & 51.8 & 71.9 & 50.6 & 54.5 & 79.7 & 94.1 & 57.2 & 63.8 \\

\cmidrule(lr){2-11}
~ & AUFM~\cite{AUFM} & 41.5 & 44.9 & 60.3 & 51.5 & 50.3 & 70.4 & 91.3 & 55.3 & 58.2 \\

~ & KDSRL~\cite{chang2022knowledge} & 60.4 & 59.2 & 67.5 & 52.7 & 51.5 & 76.1 & 91.3 & 57.7 & 64.5 \\

\cmidrule(lr){2-11}

~ & LP-Net~\cite{LP-Net} & 29.9 & 24.7 & 72.7 & 46.8 & 49.6 & 72.9 & 93.8 & 65.0 & 56.9 \\ 

~ & FAUDT~\cite{FAUDT} & 46.1 & 48.6 & 72.8 & 56.7 & 50.0 & 72.1 & 90.8 & 55.4 & 61.5 \\ 

~ & FAN-Trans~\cite{FAN-Trans} & 56.4 & 50.2 & 68.6 & 49.2 & 57.6 & 75.6 & 93.6 & 58.8 & 63.8 \\ 

\cmidrule(lr){2-11}

~ & AU-GCN~\cite{AU-GCN} & 32.3 & 19.5 & 55.7 & \textbf{57.9} & 61.4 & 62.7 & 90.9 & 60.0 & 55.0 \\

~ & SRERL~\cite{SRERL} & 45.7 & 47.8 & 59.6 & 47.1 & 45.6 & 73.5 & 84.3 & 43.6 & 55.9 \\ 

~ & UGN-B~\cite{UGN-B} & 43.3 & 48.1 & 63.4 & 49.5 & 48.2 & 72.9 & 90.8 & 59.0 & 60.0 \\

~ & HMP-PS~\cite{HMP-PS} & 38.0 & 45.9 & 65.2 & 50.9 & 50.8 & 76.0 & 93.3 & \textbf{67.6} & 61.0 \\

~ & ME-GraphAU~\cite{ME-graphAU} & 52.5 & 45.7 & \textbf{76.1} & 51.8 & 46.5 & 76.1 & 92.9 & 57.6 & 62.4 \\ 

\cmidrule(lr){2-11}

~ & \textbf{Ours} & 62.0 & \textbf{65.7} & 74.5 & 53.2 & 43.1 & 76.9 & \textbf{95.6} & 53.1 & \textbf{65.5} \\
\midrule

\multirow{5}{*}{Accuracy}  & 

EAC-Net~\cite{EAC-Net} & 85.6 & 84.9 & 79.1 & 69.1 & 88.1 & 90.0 & 80.5 & 64.8 & 80.6 \\

~ & ARL~\cite{ARL} & 92.1 & 92.7 & 88.5 & 91.6 & 95.9 & \textbf{93.9} & 97.3 & 94.3 & 93.3 \\

~ & UGN-B~\cite{UGN-B} & 95.1 & 93.2 & 88.5 & \textbf{93.2} & \textbf{96.8} & 93.4 & 94.8 & 93.8 & 93.4 \\

~ & J{\^A}A-Net~\cite{JAA-Net2020}  & \textbf{97.0} & \textbf{97.3} & 88.0 & 92.1 & 95.6 & 92.3 & 94.9 & \textbf{94.8} & 94.0 \\

\cmidrule(lr){2-11}

~ & \textbf{Ours} & 96.1 & 96.9 & \textbf{92.5} & 91.7 & 95.0 & 93.7 & \textbf{97.5} & 89.1 & \textbf{94.1}  \\

\bottomrule[1pt]
\end{tabular}
\end{center}
\end{table*}

\subsubsection{Evaluation metrics} Similar to previous methods, the F-measure (F1) score~\cite{F1score} is applied to evaluate the effectiveness of methods. The F1 score is a harmonic mean of precision and recall, with a maximum value of 1 and a minimum value of 0, which can be formulated as
\begin{equation}
    F1 = 2 \times \frac{Precision\cdot Recall}{Precision+Recall}.
\end{equation}
Besides, the accuracy is also taken into account for a more comprehensive comparison.


\subsection{Comparison with State-of-the-Art Methods}
Sixteen state-of-the-art image-based AU detection methods are compared with the proposed method, including JPML~\cite{JPML}, DRML~\cite{DRML}, EAC-Net~\cite{EAC-Net}, ARL~\cite{ARL}, J{\^A}A-Net~\cite{JAA-Net2020}, PIAP~\cite{PIAP-DF}, LP-Net~\cite{LP-Net}, FAUDT~\cite{FAUDT}, FAN-Trans~\cite{FAN-Trans}, AU-GCN~\cite{AU-GCN}, SRERL~\cite{SRERL}, UGN-B~\cite{UGN-B}, HMP-PS~\cite{HMP-PS}, ME-GraphAU~\cite{ME-graphAU}, KDSRL~\cite{chang2022knowledge} and AUFM~\cite{AUFM}. The summary and classification of each method is shown in Table \ref{tab:category}.

\subsubsection{Evaluation on BP4D} 
Table \ref{tab:BP4D result} shows the F1 score and accuracy results of different methods on BP4D. It can be seen that our method outperforms all methods based solely on region learning, including JPML, DRML, EAC-Net, ARL, J{\^A}A-Net, and PIAP, which is attributed to the capture of AU correlation information in our method. Our method also shows superiority compared to LP-Net, FAUDT and FAN-Trans, which model AU correlations without using the graph model. In particular, we compare the proposed model with graph-based methods for extracting AU correlations, including AU-GCN, SRERL, UGN-B, HMP-PS, and ME-GraphAU. As shown in Table \ref{tab:BP4D result}, our method still shows superior performance compared to other graph-based methods. The reason behind this is that the proposed SACL can mine more implicit correlation information between AUs, and the designed MSFL plays a facilitating role. Moreover, we compare our method to those that use additional data for training, namely AUFM and KDSRL. Although more relevant data (3D surface mesh or BP4D+~\cite{BP4D+}) are applied in AUFM and KDSRL, our method can obtain better results. In addition, we also compare the accuracy with popular AU detection models of EAC-Net, ARL, UGN-B and J{\^A}A-Net, and our method achieves a great improvement of 2.3\%.

\subsubsection{Evaluation on DISFA} 
Table \ref{tab:DISFA result} shows the F1 score and accuracy results of different methods on DISFA. It can be observed that our method outperforms all state-of-the-art methods by at least 1.0\% on F1 score and obtains comparable results on accuracy. This demonstrates the strong generalization ability of our method.

\begin{table*}[t]
\begin{center}
\caption{Average F1 score of different architecture implementations. For convenience, \textbf{DA} represents Data Augmentation, \textbf{$\text{Baseline}^*$} represents Baseline+DA, \textbf{FBG} represents the FACS-based graph and \textbf{SBG} represents the statistics-based graph. The best results for each column are bolded. \% is omitted.}

\label{tab:different architecture implementations}
\centering
\begin{tabular}{l p{0.6cm} p{0.6cm} p{0.6cm} p{0.6cm} p{0.6cm} p{0.6cm} p{0.6cm} p{0.6cm} p{0.6cm} p{0.6cm} p{0.6cm} p{0.6cm} p{0.65cm}}

\toprule[1pt]
 \multirow{2}{*}{Architecture}&
 \multicolumn{12}{c}{AU}  & \multirow{2}{*}{\textbf{Avg.}} \\
  
\cmidrule(lr){2-13}

~& \multicolumn{1}{c}{1} & \multicolumn{1}{c}{2} & \multicolumn{1}{c}{4} & \multicolumn{1}{c}{6} & \multicolumn{1}{c}{7} & \multicolumn{1}{c}{10} & \multicolumn{1}{c}{12} & \multicolumn{1}{c}{14} & 
\multicolumn{1}{c}{15} & \multicolumn{1}{c}{17} &
\multicolumn{1}{c}{23} & \multicolumn{1}{c}{24} &
~ \\
 \midrule
 
Baseline & 44.9 & 42.4 & 51.6 & 76.8 & \textbf{79.3} & 81.7 & 86.9 & 62.2 & 48.9 & 58.8 & 44.6 & 46.3 & 60.4 \\ 

Baseline+DA & 51.4 & 41.4 & 53.8 & 77.2 & 78.4 & 80.3 & 87.2 & 61.4 & 50.3 & 61.3 & 44.6 & 47.2 & 61.2 \\ 

\midrule

$\text{Baseline}^*$+MSFL & 55.0 & 47.7 & 57.0 & 77.6 & 79.1 & 83.4 & 88.0 & 65.2 & 52.5 & 63.1 & 47.7 & 48.9 & 63.8 \\ 

$\text{Baseline}^*$+SACL & 47.2 & 39.4 & 56.4 & 77.4 & 77.0 & 83.9 & 87.9 & 64.6 & 53.1 & 62.6 & 48.2 & 54.2 & 62.7 \\

\midrule

$\text{Baseline}^*$+MSFL+FBG & 56.9 & \textbf{50.5} & \textbf{60.0} & 77.8 & 78.9 & \textbf{84.2} & 87.8 & \textbf{66.6} & 53.3 & 64.0 & 46.8 & 48.9 & 64.6 \\ 

$\text{Baseline}^*$+MSFL+SBG & 56.0 & 49.4 & 58.9 & 77.9 & 79.2 & 84.1 & 87.7 & 65.3 & 55.9 & \textbf{64.5} & 47.9 & 49.5 & 64.7 \\ 

\midrule

$\text{Baseline}^*$+MSFL+SACL & \textbf{57.8} & 48.8 & 59.4 & \textbf{79.1} & 78.8 & 84.0 & \textbf{88.2} & 65.2 & \textbf{56.1} & 63.8 & \textbf{50.8} & \textbf{55.2} & \textbf{65.6} \\

\bottomrule[1pt]
\end{tabular}
\end{center}
\end{table*}





\begin{table}[t]
\centering
\caption{Comparison of the model complexity of our model and its variants with that of ME-GraphAU~\cite{ME-graphAU}.} 
\label{tab:model complexity}

{\begin{tabular} { l c c } 
\toprule[1pt]
Model                                   & Params~(M) $\downarrow$  & FLOPs~(G) $\downarrow$ \\
 
\midrule
$\text{Baseline}^*$                     & 10.0        & 1.47\\
$\text{Baseline}^*$+MSFL                & 12.5        & 1.66\\
$\text{Baseline}^*$+SACL                & 24.2        & 2.36\\
$\text{Baseline}^*$+MSFL+SACL (Ours)    & 27.1        & 2.57\\
ME-GraphAU~\cite{ME-graphAU}            & 94.3        & 21.33\\

\bottomrule[1pt]
\end{tabular}}
\end{table}
\subsection{Ablation Study}
We conduct ablation studies on BP4D to investigate the impact of each part of the proposed method on the results, the computational complexity of the model, also, the choice of some important parameters and settings.

\subsubsection{Impact of each part of the proposed method}

Table \ref{tab:different architecture implementations} shows the F1 score results of different architecture implementations. We set original ResT-Lite without data augmentation as the baseline. Its result is shown in the first row of Table \ref{tab:different architecture implementations}.

\textbf{Effectiveness of Data Augmentation.}
Because the collection of datasets is time-consuming and labor-intensive, the data fall into the dilemma of small numbers and limited scenarios. Data augmentation is an effective measure to increase the diversity of data. Therefore, we use random cropping, random horizontal flipping, and random color jittering for data augmentation. It can be seen from Table \ref{tab:different architecture implementations} that the introduction of data augmentation brings a 0.8\% improvement in F1 score, which indicates that richer data is indeed beneficial for model training.


\begin{figure}[t]
    \centering
    \includegraphics[width=1\linewidth]{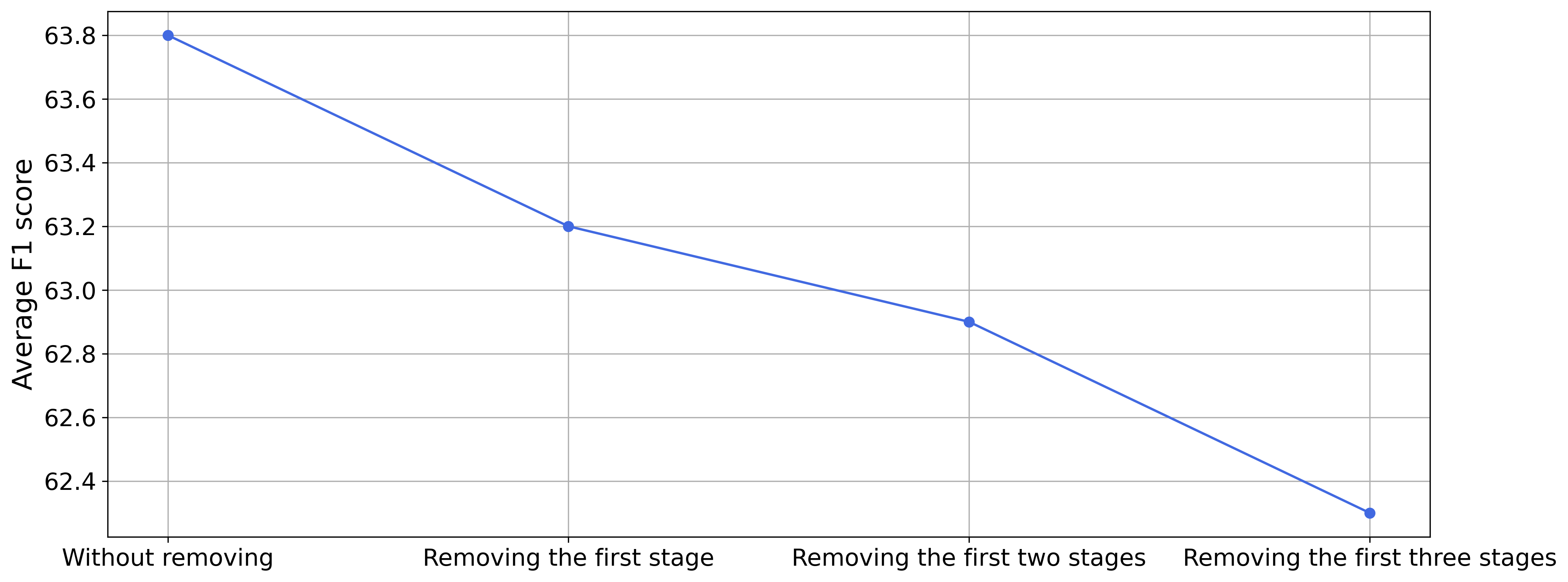}
    \caption{
    Average F1 score of removing the features from different stages in $\text{Baseline}^*$+MSFL. 
    }
    \label{concat}
\end{figure}
\begin{table*}[t]
\begin{center}
\caption{F1 score results of choosing different feature
fusion mechanisms, correlation learning ways, distance metrics and GCN types. The best results for each column in each group are bolded. \% is omitted.}

\label{tab:trysth}
\centering
\begin{tabular}{p{1.7cm} p{3.75cm} p{0.48cm} p{0.48cm} p{0.48cm} p{0.48cm} p{0.48cm} p{0.48cm} p{0.48cm} p{0.48cm} p{0.48cm} p{0.48cm} p{0.48cm} p{0.48cm} p{0.58cm}}

\toprule[1pt]
\multirow{2}{*}{Group} &
\multirow{2}{*}{Setting}&
\multicolumn{12}{c}{AU} & \multirow{2}{*}{\textbf{Avg.}}\\
  
\cmidrule(lr){3-14}

~& ~& \multicolumn{1}{c}{1} & \multicolumn{1}{c}{2} & \multicolumn{1}{c}{4} & \multicolumn{1}{c}{6} & \multicolumn{1}{c}{7} & \multicolumn{1}{c}{10} & \multicolumn{1}{c}{12} & \multicolumn{1}{c}{14} & \multicolumn{1}{c}{15} & \multicolumn{1}{c}{17} & \multicolumn{1}{c}{23} & \multicolumn{1}{c}{24} & ~ \\
\midrule
 
Fusion & Cross-attention & 55.8 & \textbf{48.8} & 58.6 & 78.6 & \textbf{79.6} & 83.9 & \textbf{88.5} & \textbf{68.1} & 55.2 & \textbf{64.2} & 49.2 & 50.4 & 65.1 \\ 

mechanism & Concatenation & \textbf{57.8} & \textbf{48.8} & \textbf{59.4} & \textbf{79.1} & 78.8 & \textbf{84.0} & 88.2 & 65.2 & \textbf{56.1} & 63.8 & \textbf{50.8} & \textbf{55.2} & \textbf{65.6} \\ 

\midrule

Correlation & Transformer~\cite{ViT} & 53.7 & 48.4 & 58.6 & 78.1 & \textbf{79.0} & \textbf{85.2} & \textbf{88.3} & \textbf{65.4} & 54.8 & \textbf{65.1} & 46.4 & 53.6 & 64.7 \\ 

learning way & GCN~\cite{Mr} & \textbf{57.8} & \textbf{48.8} & \textbf{59.4} & \textbf{79.1} & 78.8 & 84.0 & 88.2 & 65.2 & \textbf{56.1} & 63.8 & \textbf{50.8} & \textbf{55.2} & \textbf{65.6} \\ 

\midrule

Distance & Manhattan distance & 55.0 & 48.7 & \textbf{62.3} & 77.9 & 79.1 & \textbf{84.8} & \textbf{89.0} & \textbf{67.6} & 55.6 & \textbf{64.2} & 46.9 & 51.9 & 65.3 \\ 

metric & Cosine similarity & 56.0 & 48.3 & 59.4 & 78.7 & \textbf{79.6} & 83.9 & 88.1 & 64.4 & 54.0 & 62.1 & 48.4 & 54.3 & 64.8 \\ 

~ & Euclidean distance & \textbf{57.8} & \textbf{48.8} & 59.4 & \textbf{79.1} & 78.8 & 84.0 & 88.2 & 65.2 & \textbf{56.1} & 63.8 & \textbf{50.8} & \textbf{55.2} & \textbf{65.6} \\

\midrule

~ & EdgeConv~\cite{Edge} & 56.7 & 48.6 & \textbf{60.2} & 78.1 & 79.4 & \textbf{84.9} & 88.4 & 66.9 & 51.9 & \textbf{65.4} & 48.7 & 52.2 & 65.1 \\

GCN type & GIN~\cite{GIN} & 55.3 & 48.3 & 59.6 & 78.8 & 79.0 & 84.8 & 87.7 & \textbf{68.3} & 55.1 & 62.2 & 46.4 & 53.1 & 64.9 \\

~ & GraphSAGE~\cite{Sage} & 55.7 & 47.9 & 59.5 & 77.3 & \textbf{80.0} & 83.4 & \textbf{88.7} & 66.4 & 54.8 & 64.6 & 48.1 & 51.4 & 64.8 \\ 

~ & Max-Relative GraphConv~\cite{Mr} & \textbf{57.8} & \textbf{48.8} & 59.4 & \textbf{79.1} & 78.8 & 84.0 & 88.2 & 65.2 & \textbf{56.1} & 63.8 & \textbf{50.8} & \textbf{55.2} & \textbf{65.6} \\

\bottomrule[1pt]
\end{tabular}
\end{center}
\end{table*}


\textbf{Effectiveness of MSFL.}
We verify the effectiveness of MSFL, as shown in the third row of Table \ref{tab:different architecture implementations}. It can be observed that \textit{$\text{Baseline}^*$+MSFL} achieves an improvement in F1 score compared to \textit{$\text{Baseline}^*$}, which indicates that preserving spatial information and combining features of different scales obtained from different stages to enrich the receptive field is helpful for AU detection. In addition, we remove the features from the first, first two, and first three stages of MSFL respectively, and the results are shown in Figure \ref{concat}. It can be observed that the more features are removed, the poorer the receptive field of the features used for AU detection, and the lower the F1 score. This further illustrates the importance of multi-scale features in MSFL for AU detection.

\begin{figure*}[t]
    \centering
    \includegraphics[width=1\linewidth]{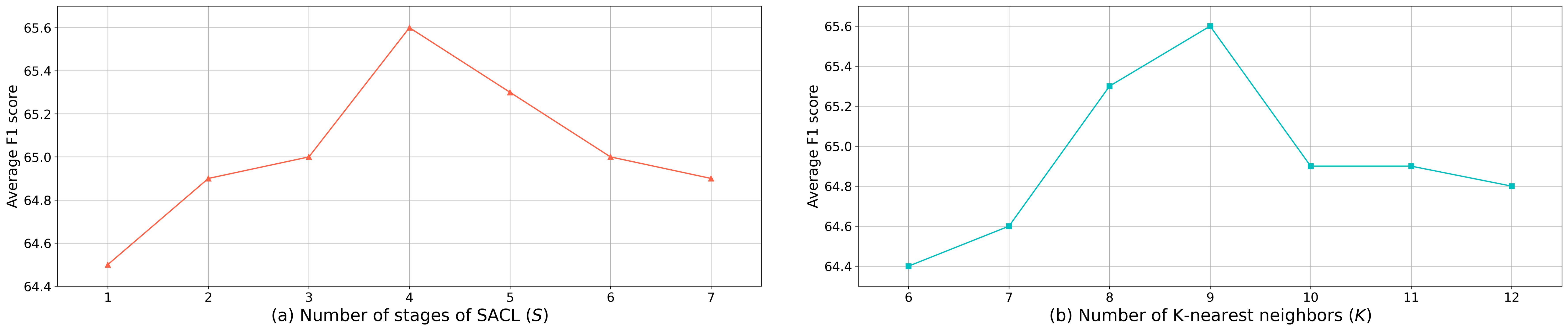}

    \caption{(a) Average F1 score of choosing different number of stages of SACL. (b) Average F1 score of choosing different number of $K$-nearest neighbors.}
    \label{k&s}
\end{figure*}


\textbf{Effectiveness of SACL.} To demonstrate the superiority of the proposed SACL, we firstly introduce it alone into \textit{$\text{Baseline}^*$} and get the variant \textit{$\text{Baseline}^*$+SACL}, which brought a 1.3\% improvement, as shown in the fourth row of Table \ref{tab:different architecture implementations}. Then, we take MSFL into account and construct several variants by introducing different graph structures. Please note that \textit{$\text{Baseline}^*$+MSFL+FBG}, \textit{$\text{Baseline}^*$+MSFL+SBG}, and \textit{$\text{Baseline}^*$+MSFL+SACL} represent the introduction of the FACS-based graph, statistics-based graph, and self-adjusting AU-correlation learning graph into \textit{$\text{Baseline}^*$+MSFL}, respectively, so the \textit{$\text{Baseline}^*$+MSFL+SACL} is the proposed method. \textit{$\text{Baseline}^*$+MSFL+FBG} adopts the AU correlation graph proposed in \cite{chang2022knowledge}, and \textit{$\text{Baseline}^*$+MSFL+SBG} adopts the AU correlation graph proposed in \cite{AU-GCN}. For fairness, the above three methods use the same graph processing. Obviously, it can be seen that the results obtained by \textit{$\text{Baseline}^*$+MSFL+FBG} and \textit{$\text{Baseline}^*$+MSFL+SBG} are better than \textit{$\text{Baseline}^*$+MSFL}, which indicates that the AU correlation information is effective for AU detection. However, both of the above are worse than the results of our \textit{$\text{Baseline}^*$+MSFL+SACL}. This is because SACL makes full use of the similarities and differences between shallow muscle motion information and deep emotion representation information by constantly updating the graph structure, thereby mining more implicit AU correlation information.



\textbf{The role of MSFL in SACL.} We explore the role of MSFL in SACL, which has rarely been mentioned in previous work. It can be observed from Table \ref{tab:different architecture implementations} that in the absence of MSFL, \textit{$\text{Baseline}^*$+SACL} brings a 1.3\% improvement compared with \textit{$\text{Baseline}^*$}. After introducing MSFL, \textit{$\text{Baseline}^*$+MSFL+SACL} can improve \textit{$\text{Baseline}^*$+MSFL} by 1.8\%. The above reveals the promotion effect of MSFL on SACL. With the introduction of MSFL, SACL can perform better. Besides, we also discuss this conclusion through the perspective of visualization, see Section \ref{sec:visualization} for details.

\subsubsection{Model complexity comparison}
In order to demonstrate the lightness and convenience of our method, we compare the model complexity with the current optimal method ME-GraphAU (based on a fully connected graph)~\cite{ME-graphAU}. Table \ref{tab:model complexity} shows the number of parameters and FLOPs of our model and its variants as well as ME-GraphAU. It can be observed that the number of parameters and FLOPs of our entire model are only 28.7\% and 12.0\% of ME-GraphAU, respectively. The reason is that ME-graphAU calculated the multi-dimensional edge features between each AU pairs ergodically based on the fully connected graph structure, which is computationally expensive. In contrast, we utilize SACL (with fewer edge connections) to adaptively learn and update AU correlation graphs by efficiently leveraging the characteristics of different levels of AU motion and emotion representation information extracted in different stages of the network, thereby mining more implicit AU correlation information with less model complexity. Importantly, the proposed network is trained in an end-to-end manner rather than a two-stage training done in ME-GraphAU.

\subsubsection{Choice of important parameters and settings}
Table \ref{tab:trysth} and Figure \ref{k&s} shows the F1 score results obtained by making different choices for important parameters and settings. Next, we will explain them one by one.

\textbf{Choice of feature fusion mechanism.}
To maximize the promotion of SACL by MSFL, it is crucial to explore the fusion method of the two. Therefore, we compare the two fusion mechanisms: cross-attention and concatenation. The implementation of cross-attention is similar to that in \cite{ME-graphAU}. The F1 score results are shown in the first group of the Table \ref{tab:trysth}. It can be seen that the effect of directly concatenating is better than applying cross-attention. We think that the reason is the feature information obtained by the two branches has different meanings, and the cross-attention may induce confusion. Conversely, concatenating the features directly could preserve the feature information to the greatest extent.


\begin{figure*}[t]
  \centering
  \includegraphics[width=1\linewidth]{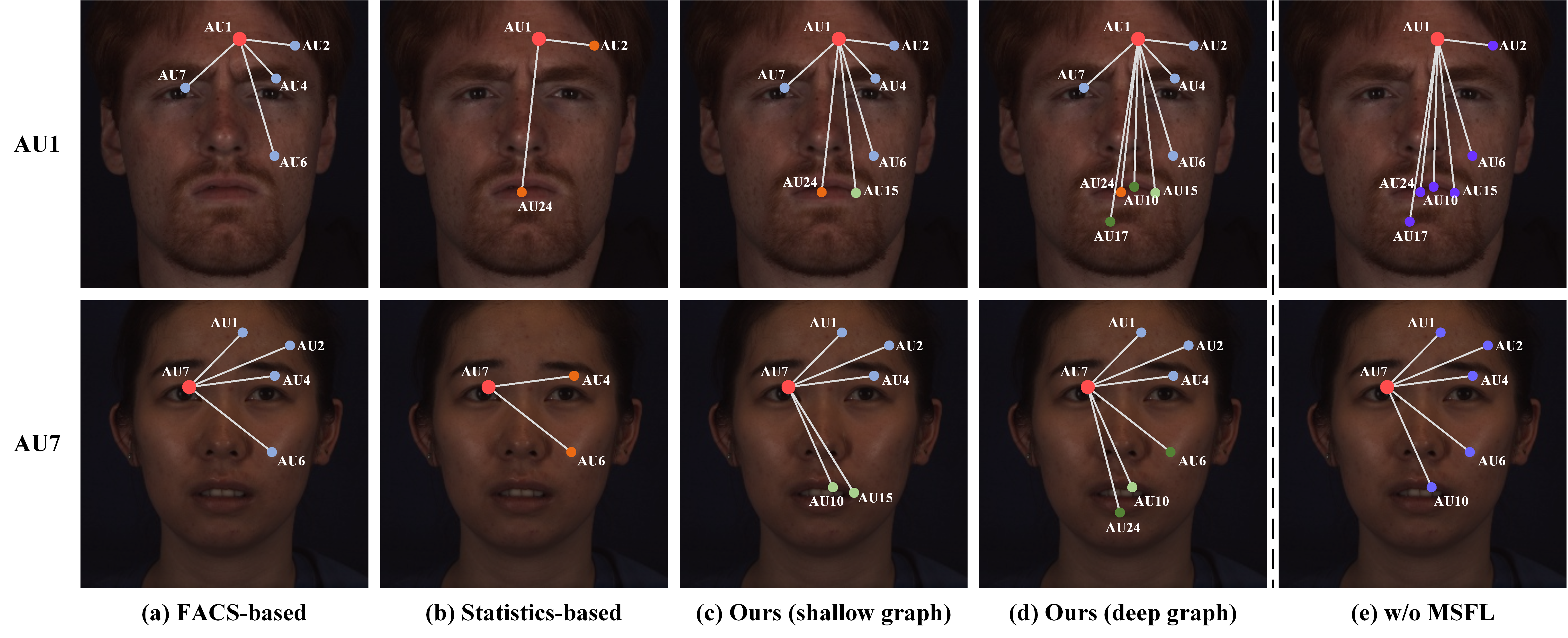}
  \caption{Visualization of AU correlation graph obtained by different methods: (a) AU correlation graph based on \textbf{FACS summary}~\cite{chang2022knowledge}; (b) AU correlation graph based on \textbf{statistical rules in the dataset}~\cite{SRERL}; (c) AU correlation graph obtained by \textbf{Shallow} layer of the proposed \textbf{SACL}; (d) AU correlation graph obtained by \textbf{Deep} layers of the proposed \textbf{SACL}; (e) AU correlation graph obtained by \textbf{Deep} layers of \textit{$\text{Baseline}^*$+SACL}. Obviously, compared with the existing methods, the proposed SACL can capture richer AU semantic correlations and better fit the actual situation, and MSFL does indeed promote the performance of SACL.
  }
  \label{visualization}
\end{figure*}

\textbf{Choice of correlation learning way.}
Due to the powerful long-distance dependency modeling capability of the Transformer~\cite{ViT}, it is also considered for correlation learning, such as in \cite{FAUDT, FAN-Trans}. We replace the GCN in the SACL with six consecutive vanilla Transformer blocks, and compare the performance of the two ways, see the second group of Table \ref{tab:trysth}. It can be observed that GCN outperforms Transformer. We suppose that this is because Transformer has large hunger on labelled data, and struggles to capture the correct correlations with AU-related small-scale datasets. In addition, the computational complexity of Transformer is higher, and just six blocks lead to 74.0M parameters and 3.41G FLOPs, which increase by 173\% and 33\% than our method, respectively. Therefore, GCN is more portable and practical for AU correlation learning.


\textbf{Choice of distance metric.} 
When updating AU correlation graph with $K$-nearest neighbor algorithm, it is worth exploring how to measure the distance between features. Therefore, we try the Manhattan distance, Cosine similarity, and Euclidean distance, as shown in the third group in Table \ref{tab:trysth}. It should be noted that we use 1 minus cosine similarity to represent cosine distance. It can be observed that the Euclidean distance achieved the best results, followed by the Manhattan distance. Therefore, when calculating the K-nearest neighbors, we choose the Euclidean distance.


\textbf{Choice of GCN type.}
The type of GCN determines the way information is aggregated and communicated between nodes. In order to choose the appropriate GCN type, we compare several representative GCN variants, including EdgeConv~\cite{Edge}, GIN~\cite{GIN}, GraphSAGE~\cite{Sage}, and Max-Relative GraphConv~\cite{Mr}. It can be observed from the fourth group of Table \ref{tab:trysth} that, Max-Relative GraphConv achieves the best average F1 score, so we choose it. In the other experiments in this paper, we use Max-Relative GraphConv by default.


\textbf{Setting of the number of stages of SACL ($S$).} To explore the optimal structure, we ablate $S$ from 1 to 7, as shown in Figure \ref{k&s} (a). When $S$ is small, the depth of network is insufficient, and the extracted features contain less deep emotion representation information of AUs, thus achieving suboptimal performance. When $S$ is large, the network tends to overfit and its performance decreases gradually. In summary, the results show that $S$ equal to 4 is the most suitable setting.


\textbf{Setting of the number $K$ in KNN algorithm.} The number $K$ of $K$-nearest neighbors is related to the construction of the AU correlation graph. As shown in Figure \ref{k&s} (b), when $K$ is equal to 9, the F1 score reaches the highest result. As $K$ gradually decreases, the F1 score decreases. This is because if too few neighbors are considered, the communication of information between nodes will be destroyed, thus losing important implicit correlation information and affecting the following construction of the AU correlation graph. Besides, when $K$ increases gradually, the F1 score also decreases. This is because if too many nodes are considered, redundant and noisy information will inevitably be introduced, which will interfere with valid information and thus make the final result worse. To summarize, we set the number $K$ in $K$-nearest neighbors to 9.

\begin{figure}[t]
    \centering
    \includegraphics[width=1.0\linewidth]{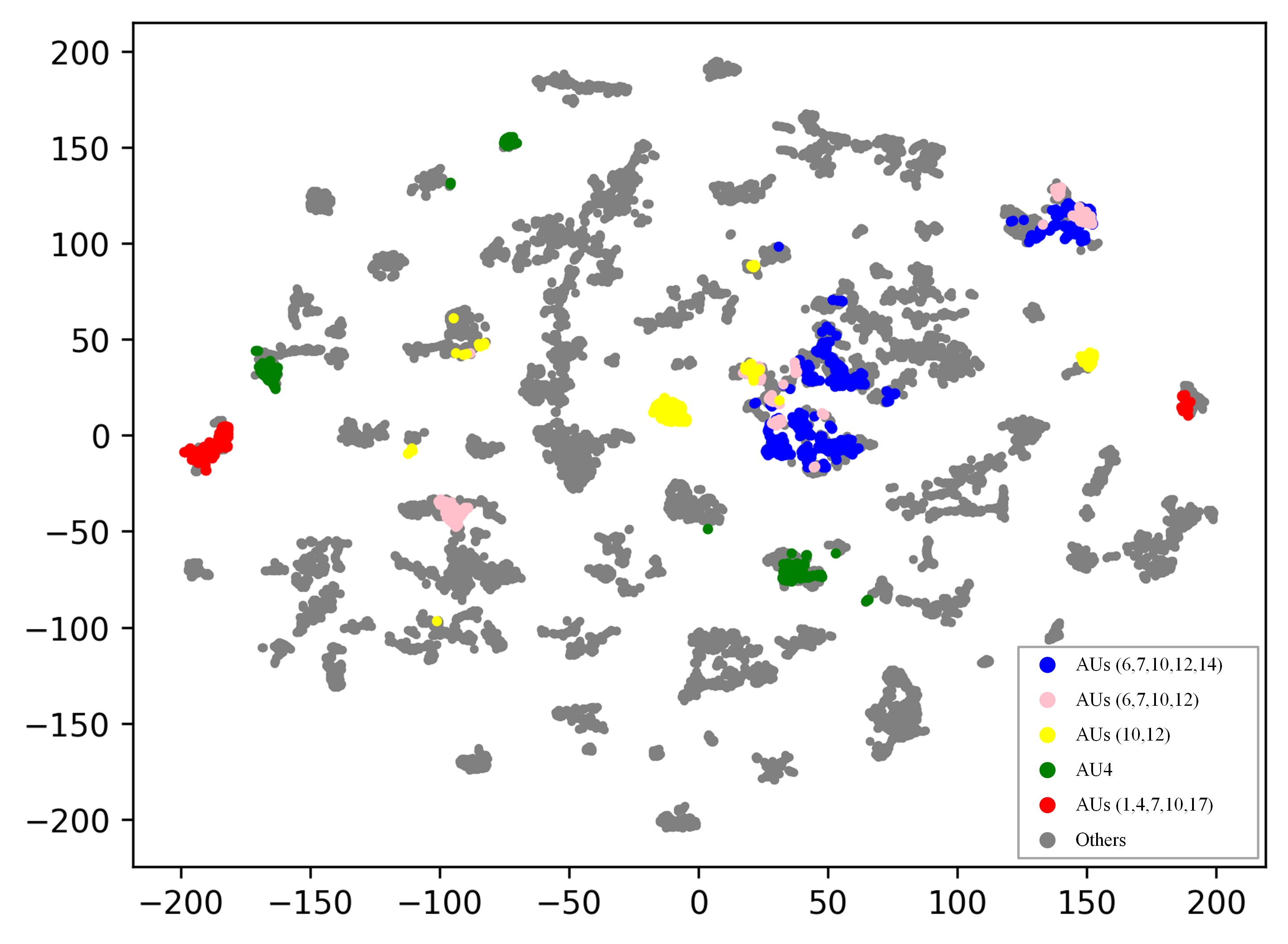}

    \caption{t-SNE distribution of our model output. We choose five common AU combinations for visualization.}
    \label{tsne}
\end{figure}

\subsection{Visualization}
\label{sec:visualization}

\subsubsection{AU correlation graph}
To better demonstrate the superiority of SACL, Figure \ref{visualization} (a)-(d) visualizes the graph structure yield by different methods; here, we utilize the AU1 and AU7 as the central node for illustration. It can be seen that the AU correlation graphs based on the FACS summary~\cite{chang2022knowledge} and the statistical rules of the dataset~\cite{SRERL} are limited and inconsistent, as shown in Figure \ref{visualization} (a) and (b). In contrast, our SACL not only considers the correlation information captured by previous methods but also obtains more implicit AU correlation information. Specifically, in the shallow layers, compared to the previous methods, more potential correlations are captured based on the similarity of the muscle motion information, such as AU1 and AU15, AU7 and AU10, as shown in Figure \ref{visualization} (c). Moreover, in the deep layers, according to the similarity of emotion representation information, we not only correct the correlations obtained in the shallow layers (such as supplementing the ignored AU7 and AU6, and removing the redundant AU7 and AU15), but also capture more implicit correlations (such as AU1 and AU17, AU7 and AU24), as shown in Figure \ref{visualization} (d). From a psychological perspective, AU1, AU15, and AU17 show a symbiotic relationship in sadness, and AU7, AU10, and AU24 always co-occur in anger~\cite{AUsurvey}, which indicates that the obtained correlation information is meaningful.

In addition, we also visualize the deep AU correlation graph obtained by \textit{$\text{Baseline}^*$+SACL}, and observe that in the absence of MSFL, the variant will miss some basic and crucial correlation information, as shown in Figure \ref{visualization} (e). This finding confirms the promoting effect of MSFL on SACL.

\subsubsection{t-SNE}

In order to demonstrate the effectiveness and discriminability of the features extracted by our method, we sample some images to visualize the final output of our model using t-SNE~\cite{t-SNE}. Figure \ref{tsne} shows the t-SNE distribution of some common AU combinations in BP4D. The legends in Figure \ref{tsne}, such as 'AUs (6,7,10,12,14)', represent that AU6, 7, 10, 12, and 14 are all activated in the sample, and other legends follow the same convention. We can observe that samples with the same AU combination are clustered closely, while those with different AU combinations are farther apart. Moreover, we found that the blue, pink, and yellow dots are closer to each other than to other dots. This is because the AU combinations of the latter two (AUs (6,7,10,12) and AUs (10,12)) are subsets of the former (AUs (6,7,10,12,14)), while the green dots with independent labels (AU4) are farther from other dots. In summary, it can be shown that the feature distribution of our model’s output makes sense.




\section{Conclusion and Future work}
\label{sec:conclusion}
This paper proposes a multi-scale promoted self-adjusting correlation learning network for AU detection. We propose a novel SACL method, which can efficiently utilize the AU motion and emotion representation information obtained in the different stages of the network to adaptively update the AU correlation graph, enabling more implicit correlation information to be mined with less computational effort. Additionally, we introduce a simple yet effective MSFL method, which has a boost to SACL. Finally, the features from SACL and MSFL are integrated to obtain a more robust representation for the final AU detection. Extensive experiments have demonstrated the superiority of our method on the widely used datasets, namely, the BP4D and DISFA.

In this paper, when the SACL module updates the AU correlation graph structure using the KNN algorithm, $K$ is set to a fixed value. In future work, gradually increasing or decreasing the value of K as the network deepens is a direction worth exploring. Besides, we will further explore the interactive effects of multi-scale learning and AU correlation learning to seek the optimal solution for information fusion between the two.





\newpage
\bibliographystyle{IEEEtran}
\bibliography{IEEEabrv,reference}

\end{document}